\documentclass[journal,10pt,doublecolumn,singlespace]{IEEEtran}
\usepackage{cite,graphicx,subfig,multirow,array}

\usepackage{amsfonts,amssymb}
\usepackage[cmex10]{amsmath}

\usepackage{algorithmic}
\usepackage[ruled]{algorithm2e}	
\usepackage{comment}

\usepackage{colortbl}
\definecolor{mygray}{gray}{.9}

\DeclareMathOperator*{\argmax}{median}

\begin{document}
\title{Haze Density Estimation via Modeling of\\Scattering Coefficients of iso-depth regions}
\author{\IEEEauthorblockN{Jie Chen, Cheen-Hau Tan, Lap-Pui Chau\\}
\IEEEauthorblockA{
School of Electrical and Electronic Engineering,\\
Nanyang Technological University, Singapore}}

	
\maketitle

\begin{abstract}

Vision based haze density estimation is of practical implications for the purpose of precaution alarm and emergency reactions toward disastrous hazy weathers. In this paper, we introduce a haze density estimation framework based on modeling of scattering coefficients of iso-depth regions. A haze density metric of Normalized Scattering Coefficient (NSC) is proposed to measure current haze density level with reference to two reference scales. Iso-depth regions are determined via superpixel segmentation. Efficient searching and matching of iso-depth units could be carried out for measurements via unstationary cameras. A robust dark SP selection method is used to produce reliable predictions for most out-door scenarios.

\end{abstract}
\begin{IEEEkeywords}
scattering coefficient, dark channel, iso-depth regions
\end{IEEEkeywords}
\IEEEpeerreviewmaketitle

\section{Background Introduction}

Extreme weather hazards happens more often these days due to climate changes and increased human industrial activities, and one of most notorious of them is haze. When haze happens, dust, smoke and other dry particles obscure the clarity of the sky, impeding human vision, causing respiratory diseases, and most dangerously, it could cause traffic and industrial accidents.

For the purpose of precaution alarm and emergency reactions for the disastrous hazy weather, haze density measurement is of practical importance. The standard solution is to use a haze meter, which measures the amount of light that is diffused or scattered when passing through a transparent material. Haze density (or usually referred to as \textit{see through quality}) is measured with a narrow angle scattering test in which light is diffused in a small range with high concentration. This test measures the clarity with which finer details can be seen through the object being tested \cite{geometries2012standard}. The haze meter also measures total transmittance. Total transmittance is the measure of the total incident light compared to the light that is actually transmitted. 

Since the haze conditions usually change fast, and the distribution of haze in a wider area is commonly uneven, a vision based haze density estimation solution is of practical implications. In this paper, we introduce a haze density estimation framework based on modeling of scattering coefficients of iso-depth regions. A density metric of Normalized Scattering Coefficient (NSC) is proposed to measure current haze density level with reference to two reference scales. Iso-depth regions are determined via superpixel segmentation. Efficient searching and matching of iso-depth units could be carried out for measurements via unstationary cameras. A robust dark SP selection method is used to produce reliable predictions for most out-door scenarios.

\subsection{The Atmospheric Model for Vision through Haze}

The foggy day image appearance is the combined result of attenuated scene radiation and transmitted airlight, whose extents both depend on the scene depth \cite{Narasimhan2002}. The classic physical model for atmospheric vision is shown in Eqn. (\ref{eqn_Trans}) :
\begin{equation}
I(x)=A\rho(x)t(x)+A(1-t(x)),\label{eqn_Trans}
\end{equation}
where $I$ is the camera captured hazy image, $A$ is the atmospheric light (referred to as airlight onwards). $\rho$ is the normalized radiance of a scene point, which is a function of the scene point reflectance. The scene albedo $J(x)=A\rho(x)$is also related to the sky illumination spectrum and intensities. $J(x)$ is independent from the weather condition, and it is the haze-free image that we aim to recover.  

The physical model in Eqn. (\ref{eqn_Trans}) describes the two physical phenomena involved when the light propagates to the camera: propagation attenuation $J(x)t(x)$, and airlight scattering $A(1-t(x))$. $t(x)$ denotes the scene point's transmission rate:
\begin{equation}\label{eqn_exponential}
t(x)=e^{-\beta d(x)},
\end{equation}
which is a exponential function related to the scene depth $d(x)$, and the \textit{scattering coefficient} $\beta$. 
The scattering coefficient $\beta$ represents the ability of a unit volume of atmosphere to scatter light in all directions \cite{Narasimhan2002}, and it directly determines the level of visual degradation the haze could cause to a given scene. 
Larger $\beta$ causes scene radiance to attenuate fast, and scatters the airlight more visibly towards the camera, which results in a contrast compromised and slightly over-exposed capture. 


\begin{figure}[t]
	\centering
	\includegraphics[width=0.9\linewidth]{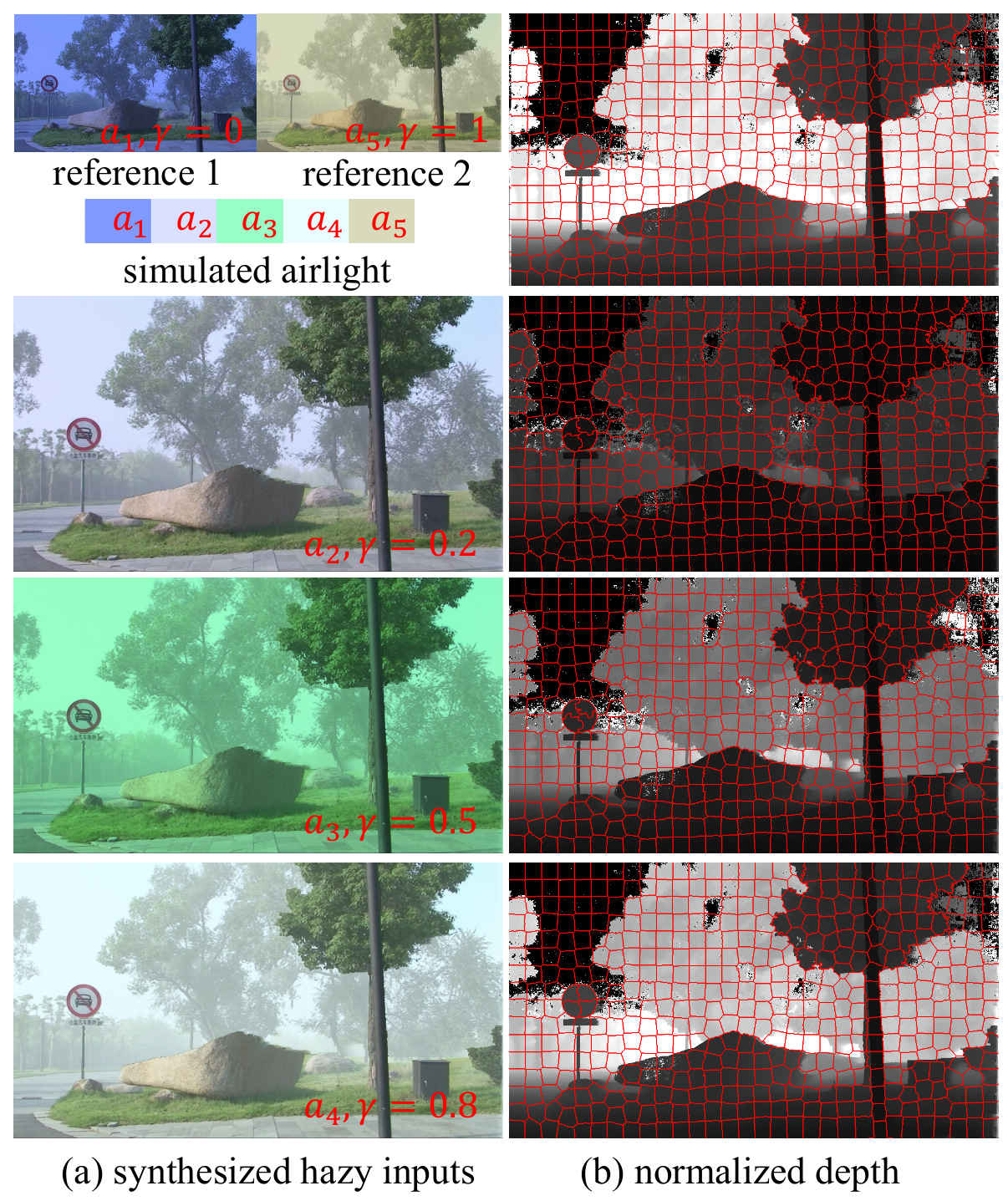}
	\caption{(a) Illustration of synthesized hazy images with different airlight, and scattering coefficient configurations. (b) Normalized depth $(\beta_t-\beta_1)d(x)$ for the scene \textit{road1} under different airlight conditions.}
	\label{fig_spOverlay}
\end{figure}

\section{Proposed Algorithm}
In this work, we propose a haze density estimation framework based on the scattering coefficient $\beta$ related to two given references.
Given two captures of a target scene captured at two different whether conditions (different scattering coefficients $\beta_1$, $\beta_2$). Suppose the camera is stationary and the scene content is static at the moment. The captured hazy images conform to the model in Eq. (\ref{eqn_Trans}):
\begin{align}
I_1(x)= A_1\rho(x) e^{-\beta_1d(x)}+ A_1(1-e^{-\beta_1d(x)}),\\ \notag
I_2(x)= A_2\rho(x) e^{-\beta_2d(x)}+ A_2(1-e^{-\beta_2d(x)}). 
\end{align}
Eliminating $\rho(x)$, we get:
\begin{equation}\label{eqn_ND}
(\beta_2-\beta_1)d(x)= -\text{ln}\frac{A_2-I_2(x)}{A_1-I_1(x)}- \text{ln}\frac{A_1}{A_2}.
\end{equation}
Based on the two captures $I_1(x)$ and $I_2(x)$, we can calculate the airlight intensities $A_1$ and $A_2$ based on the algorithms described in \cite{he2009single, chen2015heavy}. The \textbf{normalized depth} $(\beta_2-\beta_1)d(x)$ can now be directly calculated.

\subsection{Metric for Haze Density Measurement}

If given the value of $\beta_1$ and $\beta_2$ as reference haze density level, when given another arbitrary scene $I_t$ of a different weather condition (still assume that the camera is stationary and the scene content is static), we aim to estimate the relative ratio of the current $\beta_t$ with respect to the reference scattering coefficients [$\beta_1$, $\beta_2$]. We define the \textbf{normalized scattering coefficient} (\textbf{NRC}) as:
\begin{equation}\label{eqn_NRC}
\gamma= \frac{\beta_t- \beta_1}{\beta_2- \beta_1}, \gamma\in(0,1].
\end{equation}
$\gamma$ gives a direct indication how heavy the current haze is in a scale of $(0,1]$ given the two input as reference. The reference input with lighter haze makes scale "0", while the heavier one marks the scale "1". 

To estimate the NRC $\gamma$ of an arbitrary input hazy image $I_t$, according to Eq. (\ref{eqn_ND}) and (\ref{eqn_NRC}), we need to find regions between the two reference images and the target input are \textit{iso-depth}. For these iso-depth regions, depth values $d(x)$ in Eq. (\ref{eqn_ND}) could be eliminated, and the NRC can be directly calculated as:
\begin{eqnarray}
\gamma(x)=\text{ln}\frac{(I_1(x)-A_1)A_t}{(I_t(x)-A_t)A_1}~/~\text{ln}\frac{(I_1(x)-A_1)A_2}{(I_2(x)-A_2)A_1},
\end{eqnarray}
where $A_t$ is the estimated airlight for $I_t$.

It is worth noting that NRC is only a proposed indicator of this work. Any other indicator that is \textit{linearly proportionate} to the scattering coefficient can be conveniently fitted into our estimation framework.

\begin{figure}[t]
	\centering
	\includegraphics[width=3.4in]{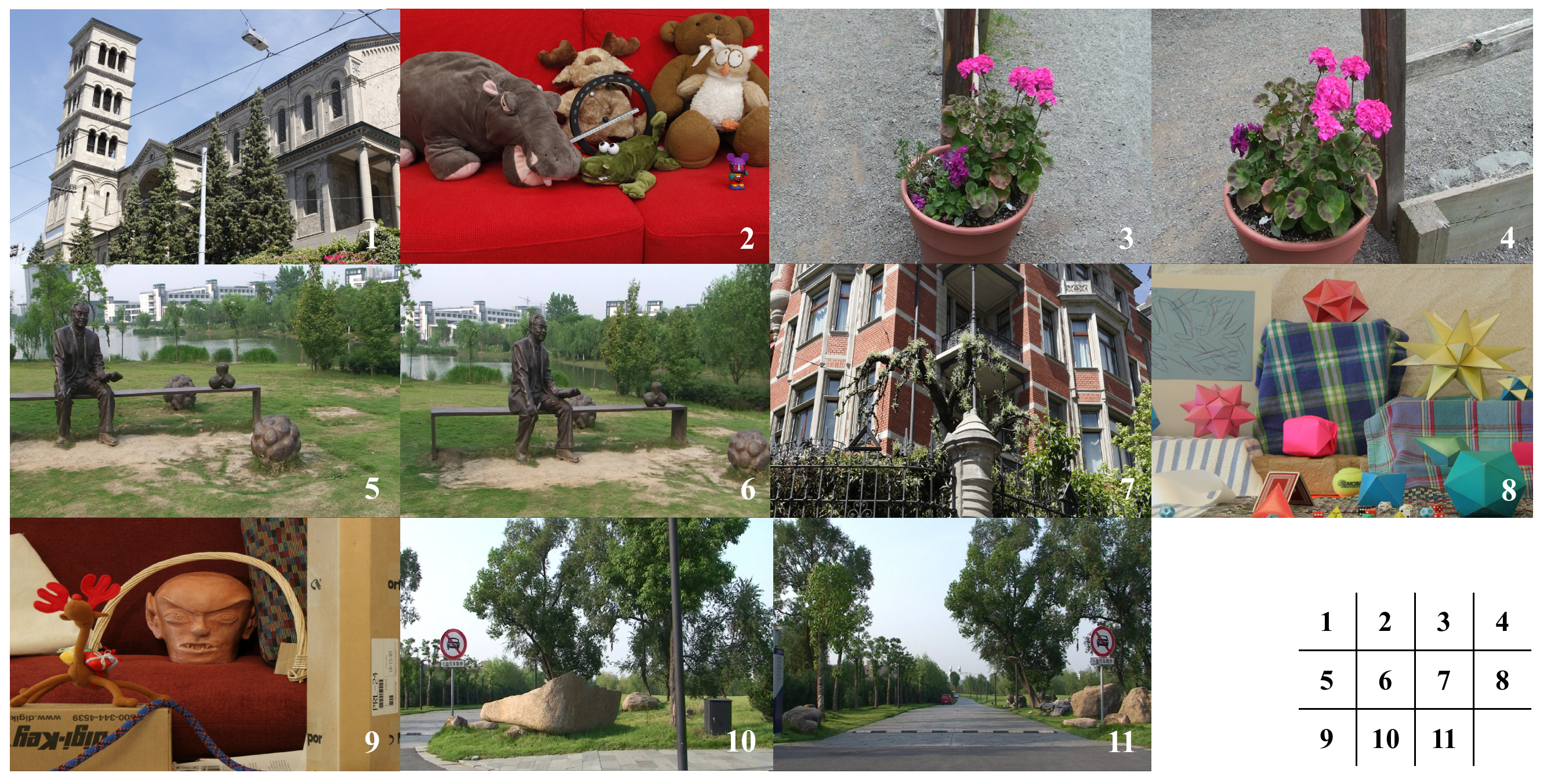}
	\caption{Haze free images from the dataset by Fattal et al. \cite{fattal2014dehazing}. These images will be used for haze synthesis with varying scattering coefficients and airlight. The scenes are 1. \textit{church}, 2. \textit{couch}, 3. \textit{flower1}, 4. \textit{flower2}, 5. \textit{lawn1}, 6. \textit{lawn2}, 7. \textit{mansion}, 8. \textit{moebius}, 9. \textit{raindeer}, 10. \textit{road1}, and 11. \textit{road2}.}
	\label{fig_dataset}
\end{figure}

\subsection{Segmentation of Iso-depth Regions}

According to Eq. (\ref{eqn_ND}), we need to find iso-depth pixels to robustly calculate the scattering coefficients. We proposed to use SuperPixel (SP) segmentation for this purpose. 
The concept of SP is to group pixels into perceptually meaningful atomic regions \cite{chen2018robust,achanta2012slic,bergh2012seeds,li2015superpixel}. Boundaries of SP usually coincide with those of the scene objects. The SPs are very adaptive in shape, and are more likely to segment uniform depth regions compared with rectangular units. 

In this work, we can assume that the pixels inside each SP are \textbf{\textit{iso-depth}}. We can estimate a robust NRC value for the SP  $\mathcal{P}_k$ based on the NRC of \text{each SP pixel} $x\in\mathcal{P}_k$: 
\begin{eqnarray}
\gamma_k=\text{median}(\gamma(x)),~x\in\mathcal{P}_k.
\end{eqnarray}
Here, $k=1,2,...,n_k$ is the SP index. Suppose there are $n_p$ SPs in total. The median filter helps to suppress noisy estimations. A NRC can be estimated for each SP.

\subsection{Robust Selection of Dark SPs}\label{sec_darkSP}
Based on the assumptions of Dark Channel Prior \cite{he2009single, chen2013enhanced}, pixels with smaller dark channel values are more likely to be close to the camera in distance. Regions with brighter dark channels (such as the sky region) tend to produce noisy and over-estimated scene depth. Therefore, we only select a set of SPs whose \textbf{median dark channel} value $\mathcal{Z}_{k}$ is below a given threshold $z_\text{n}$ to form a subset $\mathcal{S}$ from all the SPs.
\begin{equation}
\mathcal{S}= \{k'| \mathcal{Z}_{k'}\le z_\text{n},~k'=1,2,...,n_k\},
\end{equation}
The set $S$ is used to estimate the NRC for the whole image:
\begin{align}
\dot{\gamma}= \text{median}(\gamma_{k'}), k'\in\mathcal{S},\\ \notag
\end{align}

\subsection{Captures with Unstationary Caemras and Dynamic Scenes}
Until now we are holding the assumption that the cameras used to capture the reference images, and the test input image are strictly stationary, and that the scene contents are static. Such conditions can be hardly meet in practical scenarios. 

Based on the SP segmentation, we can first carry out content search and match procedure following global frame alignment as introduced in \cite{chen2018robust} and \cite{tan2014dynamic,chen2013rain}. A group of content aligned SPs between the reference and test images will be found, and all the procedures introduced above can be easily fit into a local processing framework based on these aligned SPs. As have been explained in \cite{chen2018robust}, the SP matching operation is robust to both camera motion, and scene content dynamic changes.

\section{Model Evaluation}
In this section, we evaluate the efficiency of the proposed haze density estimation framework. 

\subsection{Synthesis of Hazy Images}
First, we synthesized hazy images with image from the dataset by Fattal \textit{et al.} \cite{fattal2014dehazing}. The thumbnails for the 11 images in the dataset is shown in Fig. \ref{fig_dataset}. The dataset contains both out-door and in-door scenes, and provides ground truth transmission maps for all scenes.

To synthesize haze under different illumination conditions, we used five different airlight values with varying color tones and intensities as specified in Table \ref{tbl_airlight}. The five airlight colors ($a_1,a_2,...,a_5$) are visualized on the top-left figure in Fig. \ref{fig_spOverlay}.

\begin{table}[t]
	\begin{center}
		\small\centering\setlength\tabcolsep{1pt}
		\caption{Airlight values used for synthesis of hazy images.}
		\label{tbl_airlight}
		\begin{tabular}{|>{\centering\arraybackslash}m{1.3cm}|>{\centering\arraybackslash}m{1.3cm}|>{\centering\arraybackslash}m{1.3cm}|>{\centering\arraybackslash}m{1.3cm}|>{\centering\arraybackslash}m{1.3cm}|>{\centering\arraybackslash}m{1.3cm}|}
			\hline
			Channel &$a_1$ &$a_2$ &$a_3$ &$a_4$ &$a_5$\\\hline
			R &128 &219 &153 &234 &217 \\\hline			
			G &154 &226 &255 &253 &219  \\\hline
			B &255 &255 &198 &255 &188  \\\hline
		\end{tabular}
	\end{center}
\end{table}

Based on the haze model defined in Eq. (\ref{eqn_Trans}) and (\ref{eqn_exponential}), we set the scattering coefficients at 11 different levels: $\{0.5,0.6,0.7,...,1.4,1.5\}$, which covers the range from light haze to very heavy haze. Consequently, we have $5\times11=55$ synthesized hazy images for each input in the dataset from Fig. \ref{fig_dataset}. 
Fig. \ref{fig_spOverlay}(a) shows five different synthesis outcomes with different airlight and NRC configurations for the image \textit{road1}. Although the airlight is of different color tones, it can be observed that larger $\gamma$ values correspond to heavier haze.

We choose the parameter pair $(A_1=a_1, \beta_1=0.5, \gamma=0)$ as the synthesis parameter for the first reference input, and choose $(A_2=a_5, \beta_2=1.5, \gamma= 1.0)$ for the second reference input. The experiment results shown hereafter use the same reference settings for \textbf{all} images of the dataset \cite{fattal2014dehazing}.

Fig. \ref{fig_spOverlay}(b) shows the calculated \textbf{normalized depth maps} for the image \textit{road1} that corresponds to the hazy images on the left. As can be seen, different haze levels ($\gamma$) result in different normalized depth maps. However, the normalized depth values are similar in each SP (iso-depth regions).

\subsection{Evaluation of Dark SP Selection}

\begin{figure}[t]
	\centering
	\includegraphics[width=0.45\linewidth]{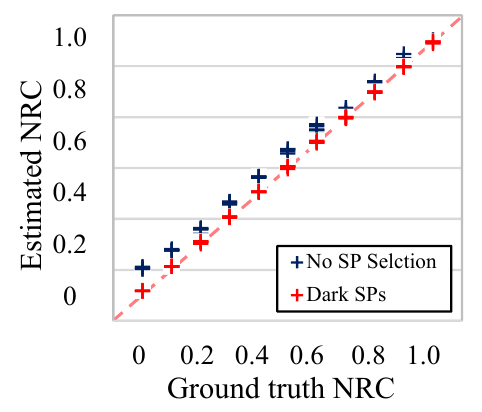}
	\caption{Comparison of $\gamma$ estimations with/without dark SP selection for the testing data \textit{church}.}
	\label{fig_darkSP}
\end{figure}

In this sub-section, we evaluate the necessity of the selection of dark SPs as explained in Sec. \ref{sec_darkSP}. We compare the estimated NRC values against the ground truth with or without the dark SP selection step on the data \textit{church}. The results are shown in Fig. \ref{fig_darkSP}. 

As can be seen, the results with dark SP selection obviously conforms better to the diagonal line, which represents the ground truth NRC value. The estimation without filtering of dark SPs over-estimates the NRC, especially for light haze conditions.

\subsection{Evaluation of Density Prediction Precision}
In this sub-section, we show the NRC estimation results against ground truth for the data \textit{church}, \textit{lawn1}, \textit{lawn2}, \textit{mansion}, \textit{road1}, and \textit{road2}. As can be observed from Fig. \ref{fig_dataset}, these data are all out-door scenes, therefore the airlight can be better estimated. The results are shown in Fig. \ref{fig_precisionPlot}(a), as can be seen, the NRC plots conform very well to the ground truth NRC values, which validates the efficiency of the proposed haze density estimation framework for \textit{out-door scenes}.

\subsection{In-door Failing Cases}
\begin{figure*}
	\begin{center}
		\includegraphics[width=1\linewidth]{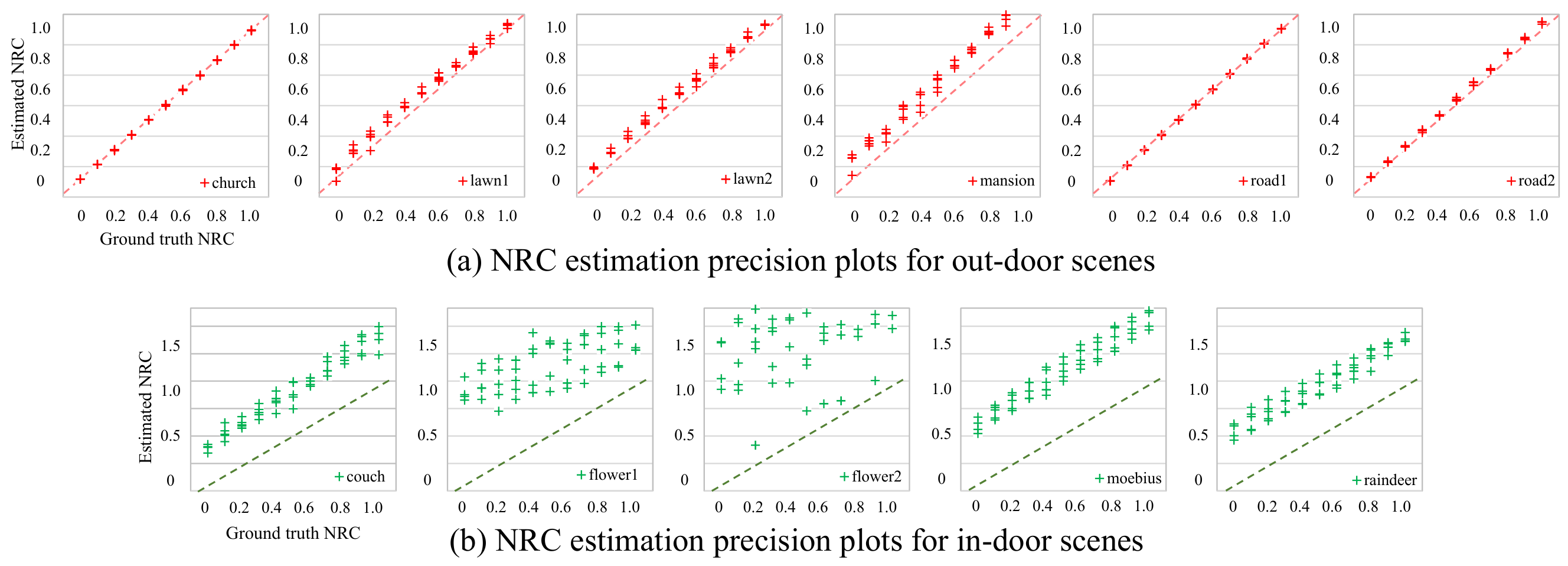}
	\end{center}
	\vspace{-0.5cm}
	\caption{Precision curves for outdoor hazy scenes.}
	\vspace{-0.4cm}
	\label{fig_precisionPlot}
\end{figure*}

For in-door scenes such as \textit{couch}, \textit{flower1}, \textit{flower2}, \textit{mobius}, and \textit{raindeer}, the prediction precision is now so good as shown in Fig. \ref{fig_precisionPlot}(b). We noticed that the limitation is mainly caused by \textbf{incorrectly under-estimated airlight values}. Since these scenes are captured in-doors, such error is understandable, since no sky regions exists. 

For these in-door cases (which we don't believe will happen in real world applications), a specialized airlight estimation algorithms can be designed to more accurately calculate the illumination source intensity, however we believe it is out of the scope of the topic discussed in this paper.

\section{Conclusions}
In this paper, we have introduced a haze density estimation framework based on modeling of scattering coefficients of iso-depth regions. 
A density metric of Normalized Scattering Coefficient (NSC) is proposed to measure current haze density level with reference to two reference scales. Iso-depth regions are determined via superpixel segmentation. 
Efficient searching and matching of iso-depth units could be carried out for measurements via unstationary cameras. A robust dark SP selection method is used to produce reliable predictions for most out-door scenarios. Any other haze density metric models that are linearly proportionate to the scattering coefficients can be easily fitted into our estimation framework.

\bibliographystyle{IEEEtran}
\bibliography{refs}

\end{document}